\documentclass[11pt]{article}

\usepackage[a4paper,margin=1in]{geometry}
\usepackage[T1]{fontenc}
\usepackage[utf8]{inputenc}
\usepackage{newtxtext,newtxmath}
\usepackage{microtype}
\usepackage{graphicx}
\usepackage{booktabs,tabularx,array,longtable}
\usepackage{amsmath,bm}
\usepackage{algorithm}
\usepackage{algpseudocode}
\usepackage{float}
\usepackage{caption}
\usepackage{placeins}
\usepackage{ragged2e}
\usepackage[
  backend=biber,
  style=authoryear,
  sorting=nyt,
  maxcitenames=2,
  maxbibnames=99
]{biblatex}
\usepackage{xurl}
\usepackage{xcolor}
\definecolor{refblue}{HTML}{1565C0}
\usepackage{hyperref}
\hypersetup{colorlinks=true, linkcolor=refblue, citecolor=refblue, urlcolor=refblue}
\usepackage{enumitem}

\newcolumntype{Y}{>{\RaggedRight\arraybackslash}X}
\newcolumntype{L}[1]{>{\RaggedRight\arraybackslash}p{#1}}
\newcolumntype{C}[1]{>{\centering\arraybackslash}p{#1}}

\title{
\textbf{Geo-SMILE: Statistical Model-Agnostic\\
Interpretability with Local Explanations for Geospatial\\
Machine Learning}
}

\author{
Shadieh Mohammadi\\
\small University of Hull\\
\small \texttt{S.MOHAMMADI-KAZAJ-2023@hull.ac.uk}
\and
Koorosh Aslansefat\\
\small University of Hull\\
\small \texttt{K.Aslansefat@hull.ac.uk}
}

\hypersetup{pdfauthor={Shadieh Mohammadi, Koorosh Aslansefat},pdftitle={Geo-SMILE: Statistical Model-Agnostic Interpretability with Local Explanations for Geospatial Machine Learning}}

\date{}

\begin{document}

\maketitle

\vspace{-0.8em}

\noindent\rule{\textwidth}{0.7pt}

\vspace{0.8em}

\noindent
\begin{tabularx}{\textwidth}{
    @{}
    L{0.30\textwidth}
    @{\hspace{0.045\textwidth}}
    Y
    @{}
}

{\small\bfseries\textls[180]{Keywords}}
&
{\small\bfseries\textls[180]{ABSTRACT}}
\\[-0.15em]

\cline{1-1}\cline{2-2}

\\[-0.25em]

\begin{minipage}[t]{\linewidth}
\small

\vspace{0.4em}

Explainable GeoAI\\
Geospatial machine learning\\
Model-agnostic interpretability\\
Geographic sensitivity\\
Spatial perturbation\\
Explanation reliability
\end{minipage}

&

\begin{minipage}[t]{\linewidth}
\small
\justifying
Conventional post-hoc explanation methods were developed largely for non-spatial inputs and often provide limited explicit treatment of geographic context. We introduce Geo-SMILE, a model-agnostic framework that distinguishes substantive Feature explanation from geographic-position response through complementary perturbation branches. The Feature branch masks substantive predictors while retaining observed coordinates; the Geo branch relocates geographic units while holding substantive predictors fixed. Following the statistical-distance-weighted surrogate principle of SMILE, input-distribution Wasserstein distances determine coalition locality weights and weighted Ridge surrogates approximate expected black-box predictions. GeoSensitivity, signed in the keep orientation, quantifies fitted-model responses to geographic-player relocation under the specified design. Across crime prediction in Hull, housing-price prediction in Seattle and ride-hailing demand in Chicago, held-out weighted surrogate fidelity ranges from $R_w^2=0.910$ to $0.989$ in the Feature branch and from $0.842$ to $0.906$ in the Geo branch. The framework makes geographic position an explicit object of post-hoc explanation while requiring separate assessment of interpretation, stability and reference dependence.
\end{minipage}

\end{tabularx}

\vspace{0.9em}

\noindent\rule{\textwidth}{0.7pt}

\vspace{1.5em}


\section{Introduction}
Geospatial artificial intelligence (GeoAI) is used to predict outcomes such as house prices, travel demand, flood risk and crime \parencite{janowicz2020,li2020}. Gradient-boosted trees \parencite{chen2016} can learn nonlinear relationships among substantive attributes and geographic coordinates, but their predictions may be difficult to interpret in planning, policing and other spatial decisions \parencite{xing2023,rousselbohm2023}. A feature ranking alone cannot show whether the fitted model responds to changes in substantive attributes or to changes in geographic position.

Spatial XAI already addresses important aspects of this problem. GLIME incorporates geographic locality into model-agnostic explanations \parencite{jin2023}, whereas GeoShapley jointly represents location and separates its attributed contribution from predictor--location interactions \parencite{li2024}. These approaches answer different questions from the response of a fitted model to the controlled relocation of geographic units. For perturbation-based explanation, that response must be defined in terms of the geographic players, their relocation rule and the reference data. It also requires evaluation: a map can appear coherent even when the coefficients are sensitive to sampling or changes in the analysis support \parencite{rousselbohm2023,krell2025}.

We introduce Geo-SMILE, a model-agnostic extension of SMILE's statistical-distance-weighted surrogate approach \parencite{aslansefat2024}. Its Feature branch masks substantive predictors while keeping coordinates fixed; its Geo branch relocates geographic players while retaining their substantive attributes. Each branch fits an interpretable linear surrogate to the response of the same black-box model, using marginal input-distribution Wasserstein distances to weight perturbation coalitions. GeoSensitivity is the Geo surrogate coefficient under the specified relocation design, reported in the keep orientation: positive values raise the expected prediction relative to relocation, negative values lower it. Feature and Geo coefficients are reported as complementary, separately estimated explanations. 
The contributions are:
\begin{itemize}
\item Introducing \textit{GeoSensitivity}, a new measure that evaluates how individual geographic units affect a model’s sensitivity to changes in location, providing a new perspective on spatial robustness and model behaviour. Unlike existing spatial explanation methods, GeoSensitivity identifies which geographic units contribute to the model’s response to geographic relocation.

\item Extending distribution-based explainability to geographic analysis by combining controlled perturbations, Wasserstein distance and weighted linear surrogates to produce interpretable feature contributions and spatial sensitivity measures, as complementary explanatory targets.

\item Demonstrating and evaluating Geo-SMILE across three geographically distinct prediction tasks, assessing explanation fidelity, consistency, spatial-support stability and correspondence with expert references.
\end{itemize}

Section~\ref{sec:literature} reviews related work, Section~\ref{sec:problem} states the explanatory questions, and Sections~\ref{sec:method}--\ref{sec:evaluation} present the method, cases and evaluation. Sections~\ref{sec:results}--\ref{sec:conclusion} report and interpret the findings.

\section{Literature review}\label{sec:literature}
We distinguish methods that attribute model predictions, methods that account explicitly for geographic context and approaches used to assess explanation reliability.

\subsection{GeoAI and the need for explainability}
GeoAI applies machine and deep learning to tasks such as land-use classification, environmental monitoring, transport modelling and urban analytics \parencite{janowicz2020,li2020}. Its observations are rarely independent: nearby places resemble one another (spatial autocorrelation), and relationships vary across space (spatial heterogeneity). Both properties matter for interpretation, since a predictor's apparent importance can be confounded with where it is observed, so explanations need to represent spatial variation rather than average it away \parencite{xing2023}. Aggregate feature-importance summaries may obscure spatial variation unless explanations are examined geographically \parencite{rousselbohm2023}. Applications to flooding \parencite{ke2025}, residential mobility \parencite{jin2023} and community satisfaction \parencite{zhang2025} show growing demand for explanations that are both sound and geographically interpretable \parencite{roussel2024}.

\subsection{Interpretable spatial models}
Earlier, interpretability was built into the model. The spatial lag model captures dependence through a lagged outcome and a weights matrix \parencite{anselin1988}; geographically weighted regression (GWR) estimates local coefficients at each location \parencite{brunsdon1996,fotheringham2002}, and multiscale GWR gives each predictor its own bandwidth \parencite{fotheringham2017}. These models are transparent but assume more regular functional forms than many GeoAI tasks support. They provide useful interpretable spatial benchmarks against which some post-hoc explanations can be compared. The Geographical Pattern Interaction model (GPI) instead uses relationships among geographic patterns and Shapley-based decomposition to estimate global and local univariate effects in interacting spatial variables \parencite{luo2025gpi}. GPI explains geographic relationships in data, whereas Geo-SMILE explains a fitted prediction model under controlled perturbations.

\subsection{Spatial XAI beyond tabular XAI}
Spatial explanation differs in three ways \parencite{rousselbohm2023,xing2023}. Location is not just another predictor: it defines neighbourhoods and context. Spatial predictors depend strongly on one another and on the coordinates, which complicates perturbation and can bias Shapley values computed under independence \parencite{aas2021}. When displayed as maps, unstable explanation coefficients may also appear to represent meaningful geographic variation.

\subsection{Progress of GeoXAI toward spatially aware methods}
\begin{figure}[htbp]\centering\includegraphics[width=\textwidth]{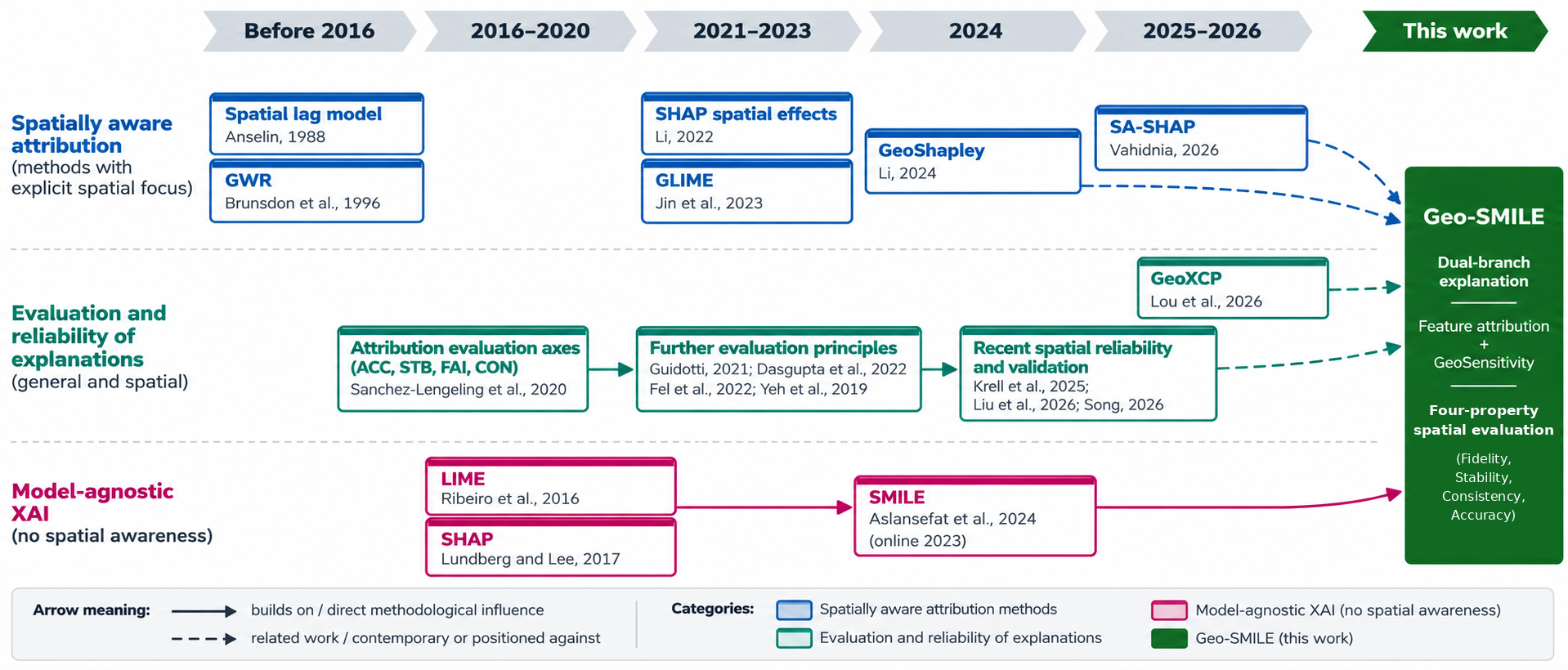}
\caption{Research strands informing Geo-SMILE; solid arrows indicate direct influence.}\label{fig:timeline}\end{figure}

Figure~\ref{fig:timeline} traces three strands: spatially aware attribution (top), the evaluation and reliability of explanations (middle) and model-agnostic XAI (bottom). Solid arrows mark direct influence, so Geo-SMILE builds on SMILE and on attribution-evaluation metrics; dashed arrows mark related work. Before 2016, explanation was embedded in spatial statistics, complemented by permutation importance and partial dependence \parencite{breiman2001,friedman2001}. LIME and SHAP \parencite{ribeiro2016,lundberg2017,lundberg2020} then made instance-level explanation routine, but treated coordinates as ordinary features. From 2021, GWR-like spatial effects were recovered from TreeSHAP values \parencite{li2022}, and GLIME redefined LIME's locality geographically \parencite{jin2023}.

GeoShapley, introduced by \textcite{li2024}, treats location as one joint player and splits each prediction into a baseline, an intrinsic location effect, location-invariant predictor effects and predictor--location interactions, analogous to spatial fixed effects and spatially varying coefficients. It has been applied to community satisfaction \parencite{zhang2025} and urban flooding \parencite{ke2025}, but its estimates depend on the chosen reference distribution and treatment of feature dependence; its location-attribution estimand differs from a controlled spatial-unit perturbation response.

Since 2025, SA-SHAP has reallocated predictor--location interactions back to the predictors and matched multiscale GWR coefficients on the Chicago tracts used here \parencite{vahidnia2026}. Attention has also shifted from what an explanation attributes to how reliable it is: GeoXCP adds conformal uncertainty intervals to attributions \parencite{lou2026}; correlated inputs can make attributions vary widely between equally accurate models \parencite{krell2025}; predictive accuracy need not match how well a model captures spatial structure \parencite{liu2026,song2026}; and geospatial deep-learning pipelines now audit attributions with deletion and stability tests \parencite{hassan2026}. Explainability has also been used to investigate what spatial structure a map-generalisation neural network learns \parencite{fu2024maps}, illustrating the role of geographic domain knowledge in interpreting black-box models.

Geo-SMILE extends perturbation-based explanation to substantive predictors and spatial units, with fidelity, stability and consistency evaluated for each branch.

\subsection{SMILE and its attribution evaluation framework}
SMILE \parencite{aslansefat2024} keeps LIME's perturbation and weighted linear surrogate but weights perturbed samples by statistical distances between the empirical distributions of the original and perturbed inputs, such as the Wasserstein, Kolmogorov--Smirnov, Kuiper and Cramér--von Mises distances, and fits the surrogate to the black box's predictions. Its authors reported closer agreement with human intuition, higher stability and greater robustness to adversarial attacks than LIME. \textcite{sanchez2020} benchmark attribution on four axes---accuracy, faithfulness, stability and consistency---scoring attribution accuracy as AUROC against a known reference, and SMILE itself quantifies explanation stability with the Jaccard index \parencite{aslansefat2024}; we refer to this family of properties as attribution (ATT) metrics. The same properties have since been reported for a generative SMILE extension \parencite{dehghani2025}. Geo-SMILE instead treats spatial units as interpretable players and evaluates both branches using the same protocol; the basic coalition-masking concept itself is inherited from LIME-style perturbation explanation and used within SMILE.

\subsection{Evaluating explanations}

Plausible explanations are not necessarily reliable. Where a reference attribution is available, explanation accuracy can be assessed directly \parencite{guidotti2021,sanchez2020}; sensitivity and infidelity provide complementary measures of how explanations respond to perturbation \parencite{yeh2019}; fidelity tests whether an interpretable surrogate reproduces the behaviour of the model it explains \parencite{dasgupta2022}; and stability is a necessary condition for reproducible interpretation \parencite{fel2022}. Feature dependence can further distort attribution estimates, particularly for Shapley-based methods \parencite{aas2021}.

For geospatial models, dependence between predictors and coordinates complicates attribution, and maps of unstable coefficients may also be mistaken for meaningful geographic patterns. We therefore assess surrogate fidelity, stability and consistency alongside reference-based attribution accuracy where an independent reference is available \parencite{mamalakis2022,krell2025}. In related spatial sensitivity analysis, \textcite{ligmannzielinska2024} combine Morris screening and Sobol variance decomposition to map how uncertain inputs influence spatial outputs. Their input-uncertainty estimand differs from Geo-SMILE's perturbation-based explanations of substantive features and geographic position in a fitted prediction model. Table~\ref{tab:methodcomparison} selects post-hoc XAI methods and explanation-reliability studies relevant to the present work. It identifies explanation scope, model dependence, explicit spatial treatment, geographic-player relocation, surrogate fitting and reliability assessment. These are methodological distinctions rather than comparative performance scores; the methods do not necessarily estimate the same quantity.

\subsection{Research gaps}
Several spatial XAI methods explicitly account for geographic context, but controlled geographic-player relocation remains a distinct explanatory target in perturbation-based surrogate methods. Estimating this response requires specifying the geographic players, relocation operator and reference distribution, together with evaluating surrogate fidelity and coefficient reproducibility.

\begin{table}[H]
\centering
\caption{XAI approaches relevant to Geo-SMILE: explanation type and methodological criteria.}
\label{tab:methodcomparison}
\begingroup
\scriptsize
\setlength{\tabcolsep}{2.5pt}
\renewcommand{\arraystretch}{1.17}
\begin{tabularx}{\textwidth}{@{}L{3.35cm}C{1.16cm}C{0.92cm}C{0.87cm}C{0.96cm}C{0.98cm}C{0.85cm}Y@{}}
\toprule
\textbf{Method} & \textbf{Ex. type} & \textbf{Class} & \textbf{Spatial} & \textbf{Reloc.} & \textbf{Surr.} & \textbf{Eval.} & \textbf{Explanation target / distinction}\\
\midrule
LIME \parencite{ribeiro2016,Molnar2025InterpretableML} & L & MA & No & No & Yes & Partial & Local feature effects via a weighted surrogate.\\
\addlinespace[2pt]
SMILE \parencite{aslansefat2024} & L & MA & No & No & Yes & Yes & Statistical-distance locality; stability and robustness.\\
\addlinespace[2pt]
SHAP / spatial TreeSHAP \parencite{lundberg2017,li2022} & L/G & MA/MS & Partial & No & No & Partial & Feature attribution and spatial patterns derived from SHAP.\\
\addlinespace[2pt]
GLIME \parencite{jin2023} & L/S & MA & Yes & No & Yes & Partial & Geographically local feature explanations.\\
\addlinespace[2pt]
GeoShapley \parencite{li2024} & L/S & MA & Yes & No & No & Partial & Joint-location, feature and interaction attributions.\\
\addlinespace[2pt]
SA-SHAP \parencite{vahidnia2026} & L/S & MA & Yes & No & No & Partial & Reallocation of predictor--location interactions.\\
\addlinespace[2pt]
GeoXCP \parencite{lou2026} & L/S & MA & Yes & No & No & Yes & Conformal uncertainty intervals for spatial explanations.\\
\addlinespace[2pt]
Reliability studies \parencite{aas2021,krell2025,hassan2026} & G/S & MA/MS & Partial & No & No & Yes & Input dependence, attribution variation and stress tests.\\
\midrule
\textbf{Geo-SMILE (this study)} & \textbf{L/G/S} & \textbf{MA} & \textbf{Yes} & \textbf{Yes} & \textbf{Yes} & \textbf{Yes} & \textbf{Feature response and geographic-player sensitivity, with separate reliability tests.}\\
\bottomrule
\end{tabularx}
\endgroup
\vspace{2pt}
\begin{minipage}{\textwidth}
\scriptsize
\emph{Notation:} L = local; G = global; S = geographically situated; MA = model-agnostic; MS = model-specific. Class = MA or MS; Spatial = explicit geographic treatment in the explanation method; Reloc. = a separate geographic-player relocation branch; Surr. = fitted interpretable approximation of black-box predictions; Eval. = explicit examination of explanation reliability or uncertainty. Yes/No indicate whether the listed method has that property; Partial indicates that it is indirect, dependent on implementation or not the method's principal focus. The criteria describe method designs, not measured superiority; the reliability measures differ across studies.
\end{minipage}
\end{table}

\FloatBarrier

\section{Problem definition}\label{sec:problem}
This study addresses three questions. (1) How can substantive-feature responses and geographic-position responses be characterised separately for a fitted geospatial predictor? (2) How should geographic-position sensitivity be defined and interpreted when geographic players are observed at different supports? (3) How faithfully and reproducibly can the resulting responses be approximated and their rankings assessed?

Geo-SMILE uses distinct Feature and Geo perturbation spaces to address the first two questions, then evaluates their weighted linear surrogates under held-out coalitions, repeated seeds, boundary changes and independent expert references where available. The resulting coefficients depend on the specified perturbations.

\section{Geo-SMILE formulation}\label{sec:method}
\begin{figure}[t]\centering
\includegraphics[width=.88\textwidth]{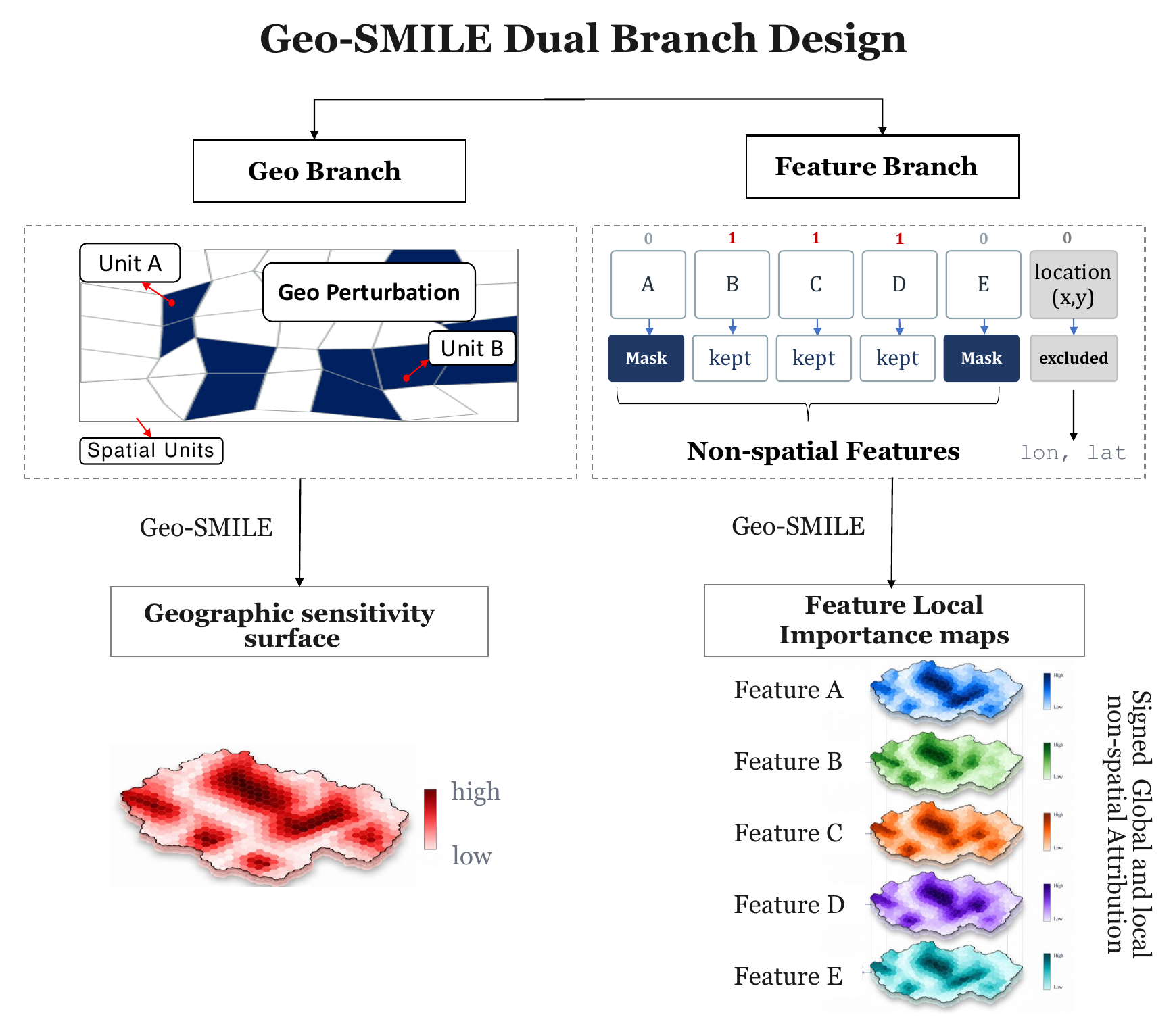}
\caption{Geo-SMILE's two explanatory targets: median-masked substantive predictors with coordinates fixed (Feature), and relocated geographic players with substantive predictors fixed (Geo).}\label{fig:dual}
\end{figure}

Geo-SMILE follows SMILE's perturb--predict--distance--weight--fit principle \parencite{aslansefat2024}, with separate perturbations for substantive predictors and geographic position. The implementation here uses binary coalitions over observation sets. The resulting coefficients explain the response of a fitted model to those specified perturbations. Figures~\ref{fig:dual} and~\ref{fig:pipeline} show the dual-branch design and workflow.

Let $X=\{(\bm x_i,\bm c_i)\}_{i=1}^{n}$, where $\bm x_i\in\mathbb R^P$ contains substantive predictors and $\bm c_i=(E_i,N_i)\in\mathbb R^2$ denotes projected coordinates. The fitted black-box regressor is $f:\mathbb R^{P+2}\to\mathbb R$. The superscript $B\in\{F,G\}$ denotes the Feature or Geo branch; $j=1,\ldots,N$ indexes a sampled coalition. 

\subsection{Perturbations}\label{subsec:dual_branches}
\paragraph{Feature branch.}\label{subsec:feature_branch}
A mask $\bm m_j\in\{0,1\}^P$ retains predictor $p$ when $m_{jp}=1$ and otherwise replaces it by its reference median $b_p$:
\begin{equation}
 x_{ip}^{(j,F)}=m_{jp}x_{ip}+(1-m_{jp})b_p,
 \qquad \bm c_i^{(j,F)}=\bm c_i.
 \label{eq:feature_operator}
\end{equation}
Masks use independent retention probability $0.5$; the all-retained mask provides the reference. Coordinates remain inputs to $f$ but are not Feature players.

\begin{figure}[htbp]
\centering
\includegraphics[width=.7\textwidth]{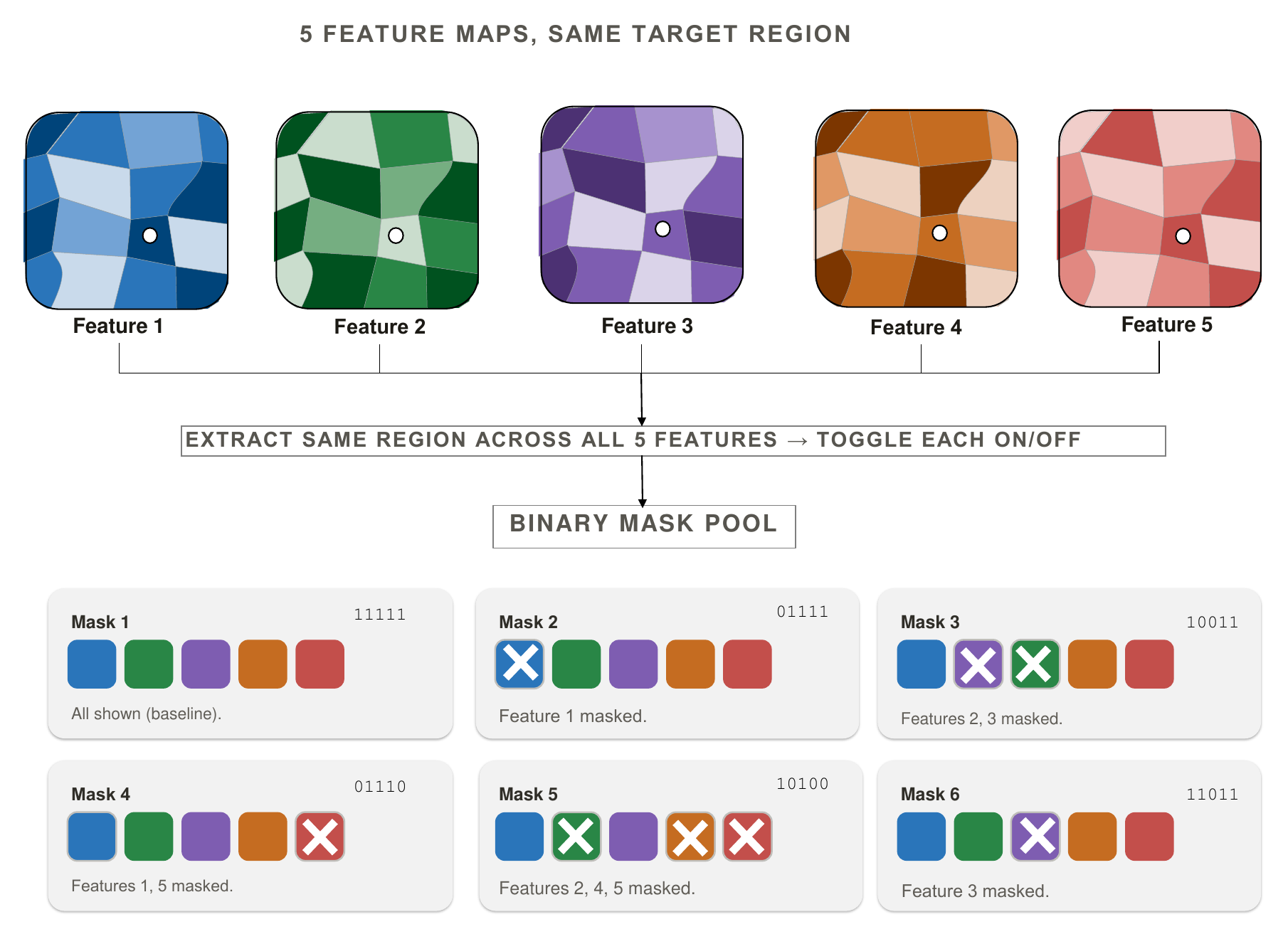}
\caption{Feature-branch masking illustrated for five predictors at one geographic unit: 1 retains the observed value, while 0 replaces it with the reference median. Coordinates remain fixed.}
\label{fig:feature_masking}
\end{figure}

Figure~\ref{fig:feature_masking} illustrates binary mask coding for one geographic unit. The same coalition mask is applied across the explained observations.

\begin{figure}[htbp]\centering
\includegraphics[width=.98\textwidth, trim=2cm 0cm 1cm 0cm ,clip]{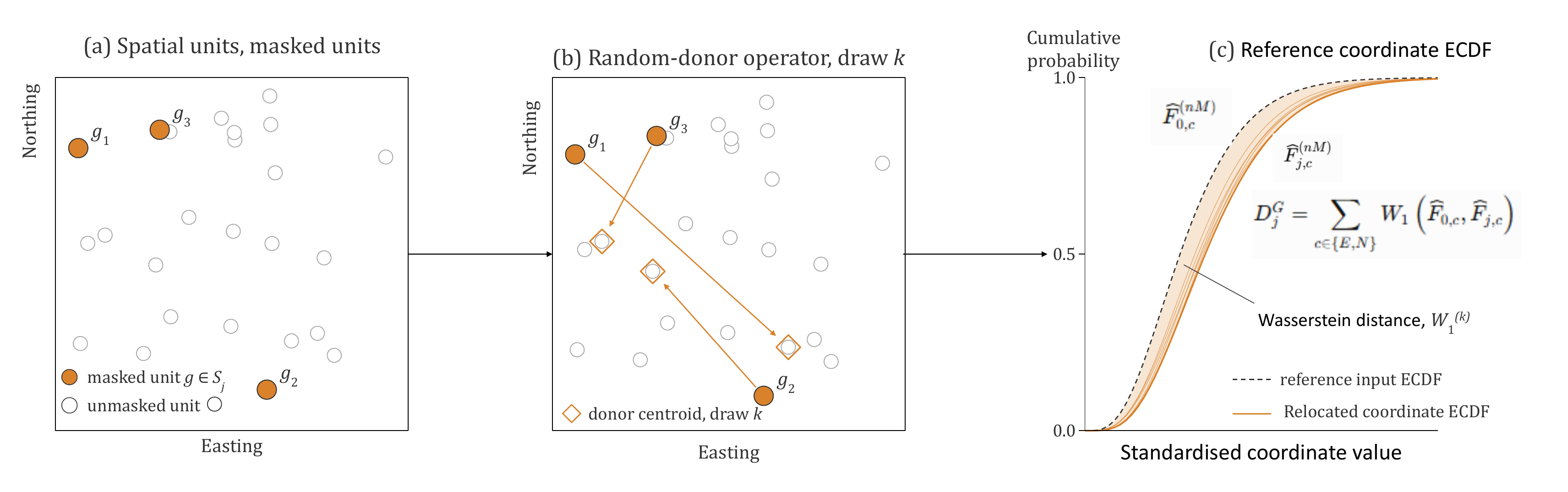}
\caption{Geographic-player relocation to randomly selected donor centroids and the input-coordinate distributions used for locality weighting. Substantive predictors are unchanged.}\label{fig:operators}
\end{figure}

\paragraph{Geo branch.}\label{subsec:geo_branch}
Let $A_g$ be the observations assigned to geographic player $g$ ($g=1,\ldots,G$). Here $s_{jg}=1$ in the geographic mask $\bm s_j\in\{0,1\}^G$ selects player $g$ for relocation. A donor $d_j(g)\ne g$ is sampled from the eligible geographic players, and the selected rows receive its representative centroid:
\begin{equation}
 \bm c_i^{(j,G)}=\begin{cases}
 \bm c_{d_j(g)},&s_{jg}=1,\\
 \bm c_i,&s_{jg}=0,
 \end{cases}
 \qquad \bm x_i^{(j,G)}=\bm x_i,\quad i\in A_g.
 \label{eq:geo_operator}
\end{equation}
The two masks therefore have different codings: Feature 1 means \emph{retain}, whereas Geo 1 means \emph{relocate}. Figure~\ref{fig:operators} illustrates relocation and its effect on the coordinate distribution. Geo coefficients refer to these players, not independently to every observation within a group. Equation~\eqref{eq:geo_operator} gives the pre-jitter relocation state; the independent Gaussian coordinate jitter described in Section~\ref{subsec:geosmile_formulation} is applied afterwards, to all $n$ observations, so an unselected player's coordinates vary across secondary samples even though $s_{jg}=0$ leaves them unrelocated.

\subsection{Prediction responses and locality weighting}
\label{subsec:geosmile_formulation}

Each coalition is evaluated over $M=10$ independent secondary samples. In the Feature branch, independent Gaussian jitter with a standard deviation equal to $0.1$ times the corresponding column standard deviation is applied to substantive predictors, including continuous, binary and ordinal variables. In the Geo branch, each secondary sample independently assigns a randomly selected eligible donor to each relocated geographic player and applies Gaussian jitter of the same relative scale to the coordinate columns of all $n$ observations. Every perturbed sample is evaluated using the same fitted black-box predictor. A sensitivity check confirms that disabling this jitter on Seattle's four binary/ordinal predictors does not materially change the fitted Feature coefficients (Section~\ref{sec:design}).

For the Feature branch, the response is retained both for observation $i$ and as a mean across observations and secondary samples:
\begin{equation}
 y_{ij}^{F}=\frac{1}{M}\sum_{k=1}^{M}f\bigl(X_i^{(j,F,k)}\bigr),\qquad
 y_j^{F}=\frac{1}{n}\sum_{i=1}^{n}y_{ij}^{F}.
 \label{eq:feature_expected}
\end{equation}
The Geo branch uses the mean prediction after donor relocation, over the same $M$ samples:
\begin{equation}
 y_j^{G}=\frac{1}{Mn}\sum_{k=1}^{M}\sum_{i=1}^{n}f\bigl(X_i^{(j,G,k)}\bigr).
 \label{eq:geo_expected}
\end{equation}

The Geo mean includes retained and relocated observations, so a player's coefficient also depends on its share of the explained observation set.

As in SMILE, statistical distance defines surrogate locality. Here the distributions are those of the \emph{inputs}, not predictions: the original and perturbed values of each permitted column are standardised using the original mean and standard deviation. If $\widehat F_{0,c}$ and $\widehat F_{j,c}$ are their empirical cumulative distribution functions, the coalition distance is
\begin{equation}
 D_j^B=\sum_{c\in\mathcal C_B}W_1\!\left(\widehat F_{0,c},\widehat F_{j,c}\right),
 \qquad \mathcal C_F=\{1,\ldots,P\},\quad \mathcal C_G=\{E,N\},
 \label{eq:marginal_input_wd}
\end{equation}
where $W_1$ is the univariate Wasserstein-1 distance between ECDFs (the integral of their absolute difference). This marginal sum does not represent a multivariate spatial distance. SMILE's exponential kernel converts it to a weight:
\begin{equation}
 w_j^B=\exp\!\left[-(D_j^B/\sigma_B)^2\right],\qquad
 \sigma_B=\operatorname{median}_{j\in\mathcal I_{\mathrm{train}}}D_j^B.
 \label{eq:smile_kernel}
\end{equation}
The bandwidth is estimated from training coalitions and then held fixed.

\subsection{Linear surrogate and GeoSensitivity}
The complex XGBoost predictor is explained by a simpler linear model. Following the compact surrogate formulation of LIME and SMILE \parencite{ribeiro2016,aslansefat2024}, each branch fits a weighted Ridge surrogate:
\begin{equation}
 \widehat h_B=\underset{h\in\mathcal H_B}{\arg\min}
 \left[\mathcal L_B(f,h,\pi_B)+\Omega(h)\right].
 \label{eq:ridge_geo}
\end{equation}
Here $\mathcal H_B$ is the class of linear surrogates $h(\bm z)=\beta_0+\bm z^\top\bm\beta$, with $\bm z_j^F=\bm m_j$ and $\bm z_j^G=\bm s_j$. The locality weight for realised coalition $j$ is $\pi_B(j)=w_j^B$ from Equation~\eqref{eq:smile_kernel}, and
$\mathcal L_B(f,h,\pi_B)=\sum_{j\in\mathcal I_{\mathrm{fit}}} w_j^B\bigl(y_j^B-h(\bm z_j^B)\bigr)^2$
is its weighted squared prediction error. The Ridge penalty is $\Omega(h)=\alpha\lVert\bm\beta\rVert_2^2$ ($\alpha=1$); the intercept is unpenalised. The fitted coefficients of $\widehat h_B$ are denoted $\widehat{\bm\beta}^{B}$. For local Feature maps, $y_{ij}^F$ replaces the mean response. Feature coefficients describe retention relative to median masking, whereas Geo coefficients describe relocation under the donor design. Since the Geo mask codes relocation as 1, we report GeoSensitivity in the keep orientation,
\begin{equation}
 \operatorname{GeoSensitivity}_g=-\widehat\beta_g^G,
 \label{eq:geosensitivity}
\end{equation}
the coefficient of keeping player $g$ at its true location rather than relocating it to a random donor: a positive value raises the expected prediction relative to relocation, a negative value lowers it. Its absolute value, $|\operatorname{GeoSensitivity}_g|$, discards this direction (Section~\ref{sec:results}, Figure~\ref{fig:geosurf}). Both views are used in this paper: the signed value for direction and for the ATT consistency and stability full-vector comparisons, and the magnitude for the GeoSensitivity surfaces, the boundary-stability top-20 overlap (Table~\ref{tab:stability}) and the Hull expert-reference comparison (Table~\ref{tab:hull_att_accuracy_primary}). GeoSensitivity quantifies the fitted model's sensitivity to specified geographic relocations, conditional on the selected players, donor pool and observation support; it does not establish causal effects of location.

\begin{figure}[htbp]\centering
\includegraphics[width=.92\textwidth, trim=0cm 1.22cm 0cm 1cm, clip]{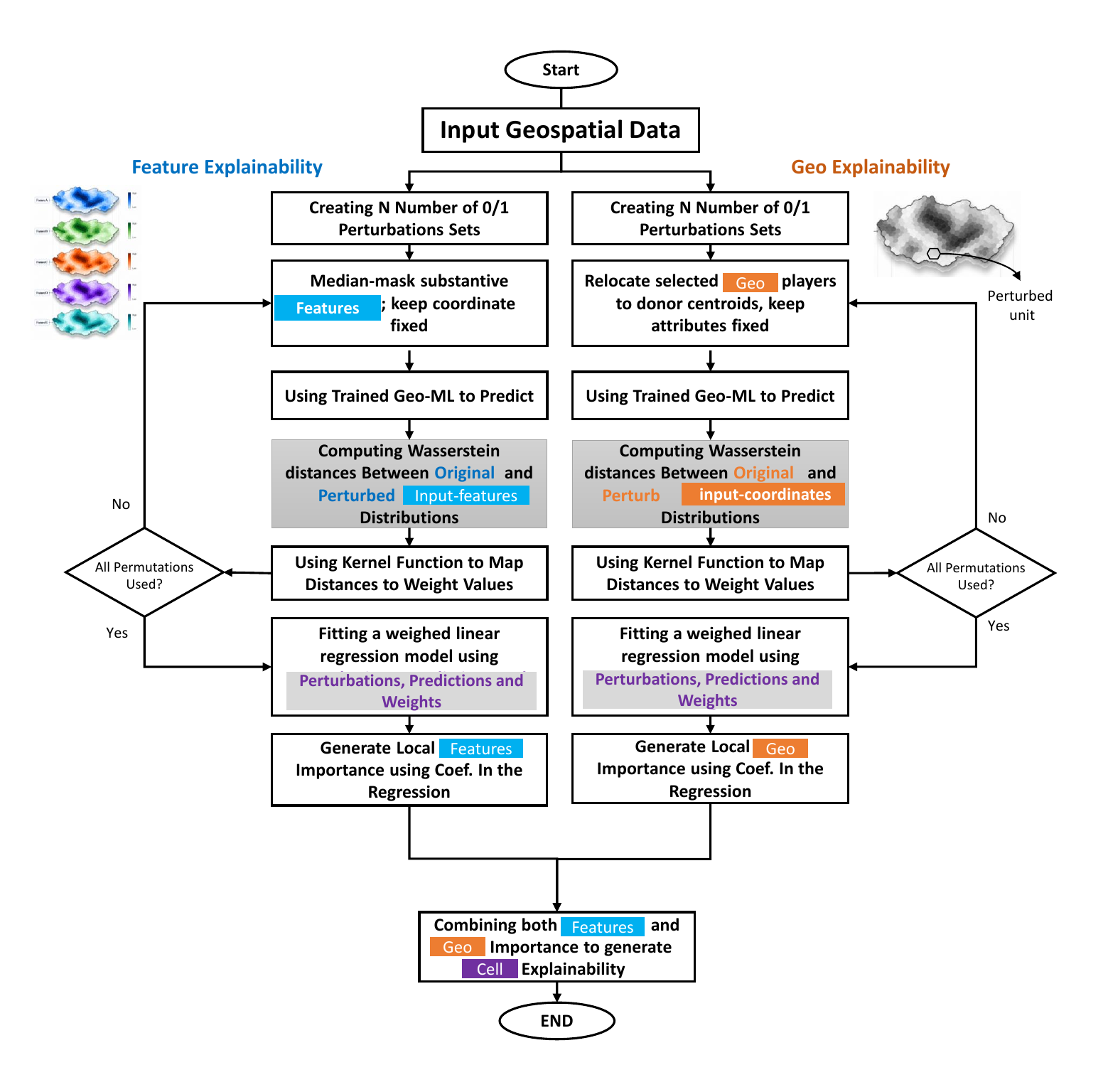}
\caption{Geo-SMILE workflow from separate Feature and Geo perturbations to surrogate outputs and their joint presentation in the six-role Cell Explainability dashboard. The branch coefficients are not added.}\label{fig:pipeline}
\end{figure}

\paragraph{Implementation.}
For each branch, $N=4{,}000$ coalition masks (seed 42) are partitioned 60/20/20 into training, validation and test sets by \emph{distinct mask}. The kernel bandwidth uses training distances. Held-out fidelity comes from a surrogate fitted on training coalitions; reported coefficients are refitted on training and validation coalitions without using the test masks. Each Geo coalition selects a coalition size drawn uniformly at random from a fixed integer range and that many geographic players, chosen uniformly at random without replacement, for relocation: 5--10 of Hull's 168 LSOAs, 5--10 of Seattle's 100 groups, and 8--16 of Chicago's 160 groups (Section~\ref{sec:cases}). The complementary branch outputs are presented in the six-role Cell Explainability dashboard (Figure~\ref{fig:pipeline}).

\FloatBarrier

\section{Geospatial cases}\label{sec:cases}
The three cases differ in outcome, spatial support and how much location matters (Tables~\ref{tab:cases} and~\ref{tab:players}). Seattle and Chicago reuse the data and models from the corresponding spatial-XAI studies \parencite{li2024,li2022}; Hull is an independent application with an expert reference.

\begin{table}[htbp]\centering
\caption{The three geospatial cases.}\label{tab:cases}
\begin{tabularx}{\textwidth}{@{}p{3cm}YYY@{}}\toprule
& Hull & Seattle & Chicago\\\midrule
Outcome & Next-month crime count & $\log_{10}$ sale price & log ride-hailing trips\\
Units & 168 LSOAs, monthly & 16,581 sales (points) & 795 census tracts\\
Predictors & 30 (crime history, census, built environment) & 8 (structure, quality) & 9 (socio-economic, built environment)\\
Coordinates & LSOA centroid (BNG) & Sale location & Tract centroid\\
Model selection & Top tier of ten screened models & Best of six models reported by \textcite{li2024} ($R^2=0.908$) & Model used by \textcite{li2022}\\
Model; test $R^2$ & XGBoost (Poisson); 0.895 & XGBoost; 0.894 & XGBoost; 0.794\\
Test MAE & 4.83 crimes per month & 0.054 (log10 price) & 0.603 (log trips)\\
Held-out units & Months from January 2026 & 3,317 sales & 159 tracts\\
Relocation operator & \multicolumn{3}{c}{Random donor centroids; one assignment per selected player}\\
Stability test & Real neighbour added & Boundary hold-out & Boundary hold-out\\
Accuracy reference & Expert predictor list and map & -- & --\\\bottomrule
\end{tabularx}\end{table}

\subsection{Hull: next-month crime prediction}
Hull is the one case with an independent expert reference.

Data and temporal design. The unit is the LSOA-month. Police-recorded street crime \parencite{homeoffice2026} for Hull's 168 LSOAs forms a complete panel, with zeros for months without offences; the outcome is next month's crime count. The 30 predictors (Table~\ref{tab:players}) combine crime history computed within each LSOA, seasonality, 2021 Census variables \parencite{ons2022a} and OpenStreetMap built-environment measures \parencite{osm2026}, plus the LSOA centroid coordinates \parencite{ons2022b}. The split is chronological: training to June 2025, validation July--December 2025, testing from January 2026.

Expert reference. A domain expert identified ten predictors considered most relevant to next-month crime and delineated areas of Hull considered particularly informative for crime patterns (Figure~\ref{fig:expert}). These were fixed in advance as the reference for ATT accuracy. An LSOA belongs to the spatial reference if at least 20\% of its area, or $0.1\,\mathrm{km}^2$, lies inside an expert area, giving 58 of 168 LSOAs. Both reference sets were specified by the same expert.

Model screening. Ten candidates from three families (tree ensembles, kernel methods, neural networks) were screened on predictive accuracy and on surrogate fidelity in both branches. The Gaussian process, XGBoost, CatBoost and LightGBM form a top tier within 0.006 in predictive $R^2$ , whereas a linear surrogate could not capture the Feature responses of two neural models (Keras and MLP). Below the top tier come ANN (0.878), SVR (0.868), CNN (0.861), Keras (0.788), Transformer (0.753) and MLP (0.527). Adding coordinates barely changes accuracy (XGBoost 0.8952 with, 0.8957 without). XGBoost ($R^2=0.895$, MAE = 4.83, RMSE = 6.91) is retained as a top-tier member that matches the model family used in Seattle and Chicago.

\subsection{Seattle: house prices}

The Seattle case uses the King County house-sales data analysed by \textcite{li2024},
originally sourced from the GeoDa Lab \textit{Home Sales} dataset
\parencite{geoda_kchomesale}. After preprocessing, 16,581 sales remain, with eight
housing predictors, projected sale coordinates (UTM zone 10N), and $\log_{10}$ sale price as the
target. XGBoost was the best of six models reported by \textcite{li2024}. Before fitting the Geo surrogate, KMeans partitions the sale coordinates into $G=100$ spatial groups, each treated as one Geo player; the 100 Ridge coefficients are group-level quantities and are not independent house-level effects. Seattle adopts whole-region random donors and the expected-prediction Geo response (Equation~\eqref{eq:geo_expected}).

\subsection{Chicago: ride-hailing demand}

The Chicago case follows the spatial-effects study by \textcite{li2022}, using ride-hailing trips from the City of
Chicago Transportation Network Providers database \parencite{chicago_tnp_trips},
socioeconomic variables from the American Community Survey
\parencite{us_census_acs}, and built-environment indicators from the EPA Smart
Location Database \parencite{epa_smartlocation}. The final data contain 795 census
tracts, nine predictors, projected tract-centroid coordinates (UTM zone 16N; EPSG:32616) and a log-transformed trip-demand target. The 795 tracts are grouped into 160 KMeans geographic players for the Geo-branch surrogate. Random-donor relocation changes the locations of selected players while holding substantive attributes fixed.

\begin{table}[htbp]\centering
\caption{Feature, location and geographic players in the three cases.}\label{tab:players}
\begin{tabularx}{\textwidth}{@{}p{1.7cm}YYY@{}}\toprule
& Hull (30 predictors) & Seattle (8 predictors) & Chicago (9 predictors)\\\midrule
Feature players & Crime history (one-, two- and three-month lags; three- and six-month rolling means); seasonality; demography; tenure; labour, health and education; built environment & Living area; lot size; construction grade; condition; view; waterfront; age; bathrooms & Mean trip distance; \% bachelor's degree or higher; \% aged 18--34; \% shared trips; \% white; \% households without a vehicle; log population density; job entropy; road-network density\\
Location inputs & LSOA centroid easting and northing (BNG) & Sale coordinates (UTM zone 10N) & Tract centroid coordinates (UTM zone 16N)\\
Geo players & 168 LSOAs; 5--10 relocated per coalition & 100 KMeans groups of sales & 160 KMeans groups of 795 tracts\\\bottomrule
\end{tabularx}\end{table}

\subsection{Explanation design}
Each case explains one XGBoost model fitted on the predictors and coordinates; in Hull it is a Poisson count model (500 trees, learning rate 0.04, depth 5, minimum child weight 3, subsampling 0.85, L2 penalty 2.0). All Geo branches use whole-region random donors, input-Wasserstein locality weights and black-box prediction targets averaged over the explained observations. Geo coalitions relocate selected spatial units (5--10 LSOAs in Hull); the geographic support has 168 Hull LSOAs, 100 Seattle groups and 160 Chicago groups. All branches draw $N=4{,}000$ coalitions; in the lower-dimensional Feature branches some binary masks can recur, but repeated masks are assigned to the same data partition; kernels are selected using a 60/20/20 training, validation and test split, Ridge uses $\alpha=1$, and seed 42 gives the reported explanations. Consistency is assessed across five seeds including seed 42.

\section{Geo-SMILE evaluation framework}\label{sec:evaluation}
Geo-SMILE is evaluated with four attribution (ATT) properties, adapted for the spatial setting from the attribution-benchmarking axes of \textcite{sanchez2020} and SMILE's stability analysis \parencite{aslansefat2024}: fidelity, stability, consistency and accuracy. Fidelity, stability and consistency are assessed in all three cases, while ATT accuracy is evaluated in Hull against an independent expert reference. Two additional diagnostics examine perturbation budget and design choices. Figure~\ref{fig:evaluation} and Table~\ref{tab:evaluation} summarise the framework. The ROC panel is schematic and illustrates how reference labels and explanation scores are used to calculate AUROC.

\subsection{Primary ATT evaluation}\label{subsec:primary_att}
\begin{figure}[htbp]\centering\includegraphics[width=.98\textwidth, trim=2.5cm 1cm 3cm 0.5cm , clip]{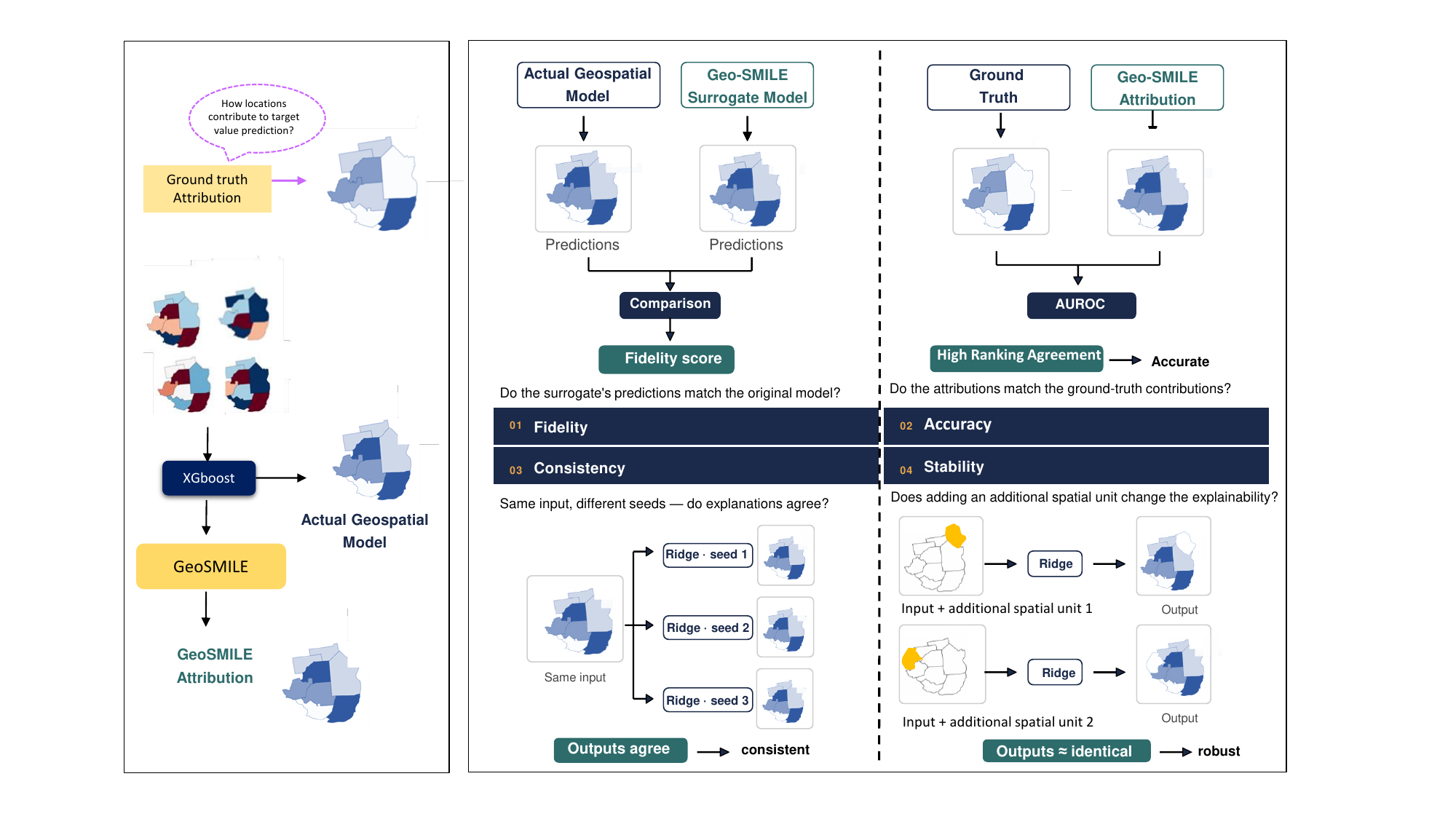}
\caption{Four dimensions of explanation evaluation: surrogate fidelity, spatial-support stability, consistency and reference-based attribution accuracy (Hull only).}\label{fig:evaluation}\end{figure}

Figure~\ref{fig:evaluation} illustrates the four dimensions used to evaluate Geo-SMILE explanations. Together, these assess how closely the surrogates reproduce black-box predictions, how reproducible the explanations are, how they respond to changes in spatial support, and their agreement with an independent expert reference where available.

\subsubsection{ATT fidelity}
Held-out surrogate fidelity measures how closely each branch's weighted linear surrogate reproduces the black-box prediction target for unseen perturbation coalitions. We report the coefficient of determination $R^2$, weighted $R_w^2$, weighted mean absolute error (WMAE) and weighted root mean squared error (WRMSE). The evaluation uses coalition-disjoint partitions to assess generalisation beyond the masks used for surrogate fitting.

\subsubsection{ATT stability}
The natural small change for a spatial explanation is to the study area. One unit is added at each cardinal direction in turn, with everything else fixed, and rankings are compared by top-$k$ Jaccard $J(A,B)=|A\cap B|/|A\cup B|$, full-ranking Spearman correlation and the added unit's rank. Hull adds a real neighbouring LSOA. Seattle and Chicago use boundary hold-out protocols to assess changes under an expanded spatial support. Seattle uses four quiet-anchor hold-outs, one in each cardinal direction. For each test, a cluster of 166 sales is withheld, XGBoost is retrained, and the cluster is reintroduced as one additional geographic group (101 players). Chicago similarly uses four quiet-anchor hold-outs (north, south, east and west): five tracts are withheld in each run, XGBoost is retrained, and the held-out cluster is reintroduced as one Geo player (161 players). 

\subsubsection{ATT consistency}
Because coalitions (and donors) are random, each branch is recomputed across five seeds with other modelling choices fixed. The primary consistency measure is each coefficient's across-run variance,
\begin{equation}
s_j^2=\frac{1}{R-1}\sum_{r=1}^{R}\left(\beta_j^{(r)}-\bar\beta_j\right)^2,
\end{equation}
reported as its standard deviation $s_j$ and summarised over players within each branch by the mean and maximum; only the mean and maximum SD are reported in this paper.

\subsubsection{ATT accuracy}
Attribution accuracy is evaluated in Hull against an independent expert-defined reference: ten substantive predictors for Feature and 58 of the 168 LSOAs for Geo. This provides an external benchmark where ground-truth model attributions are unavailable. Following \textcite{sanchez2020}, we report ATT ACC, F1 and AUROC, together with top-$k$ overlap against the chance level $|\mathcal R|/G$ (lift), where $\mathcal R$ denotes the expert-highlighted geographic units. A one-sided hypergeometric test gives
\begin{equation}
p=\Pr(X\ge m),\qquad X\sim\mathrm{Hypergeometric}(G,|\mathcal R|,k).
\label{eq:hypergeom}
\end{equation}
The hypergeometric reference assumes exchangeable spatial units; its $p$-values are reported without adjustment for spatial autocorrelation.

\FloatBarrier

\begin{table}[htbp!]
\centering
\caption{Geo-SMILE explanation evaluation and diagnostic measures.}
\label{tab:evaluation}
\begin{tabularx}{\textwidth}{@{}p{3cm}Yp{6.2cm}@{}}
\toprule
Category  & Evaluation question & Measures \\
\midrule

\multicolumn{3}{@{}l}{\textbf{Primary ATT evaluation}} \\[0.3em]

ATT fidelity &
Does the surrogate reproduce the black-box perturbation response on unseen coalitions? &
Held-out $R^2$ and weighted $R^2$; WMAE; WRMSE. \\

ATT stability &
Does the explanation remain similar after a small change to the spatial study area? &
Top-$k$ Jaccard; full-ranking Spearman correlation; rank assigned to the added or re-admitted spatial unit. \\

ATT consistency &
Do independently seeded runs produce reproducible attribution scores? &
Across-run coefficient standard deviation: mean and maximum over players. \\

ATT accuracy &
Do the resulting rankings correspond with an independent attribution reference where one is available? &
ATT ACC; F1; AUROC; top-$k$ overlap; lift above chance; hypergeometric $p$-value; single-expert correspondence. \\

\midrule
\multicolumn{3}{@{}l}{\textbf{Complementary validation diagnostics}} \\[0.3em]

Perturbation budget &
Is the coalition budget sufficient for the number of explanatory players being estimated? &
Held-out surrogate fidelity across increasing perturbation budgets $N$; convergence and plateau behaviour. \\

Design diagnostics &
Are the main conclusions robust to methodological configuration choices? &
Surrogate-family comparison and the Seattle categorical-jitter check. \\

\bottomrule
\end{tabularx}
\end{table}
\FloatBarrier

\subsection{Complementary diagnostic design}
Two diagnostics supplement the primary ATT evaluation. Perturbation-budget analysis examines held-out weighted surrogate fidelity over $N=1{,}000$--$10{,}000$ coalitions in all three cases. The surrogate-family analysis compares linear estimators under the stated Feature and Geo configurations.

\section{Results}\label{sec:results}

\subsection{Feature attributions}
Global Feature coefficients quantify fitted conditional associations between retaining a substantive predictor and the expected black-box prediction under median masking, with coordinates held fixed. Their signs show the direction of this model-response contrast. Local coefficients reveal how the contrast varies across geographic observations.

\begin{figure}[htbp]\centering
\includegraphics[width=.98\textwidth]{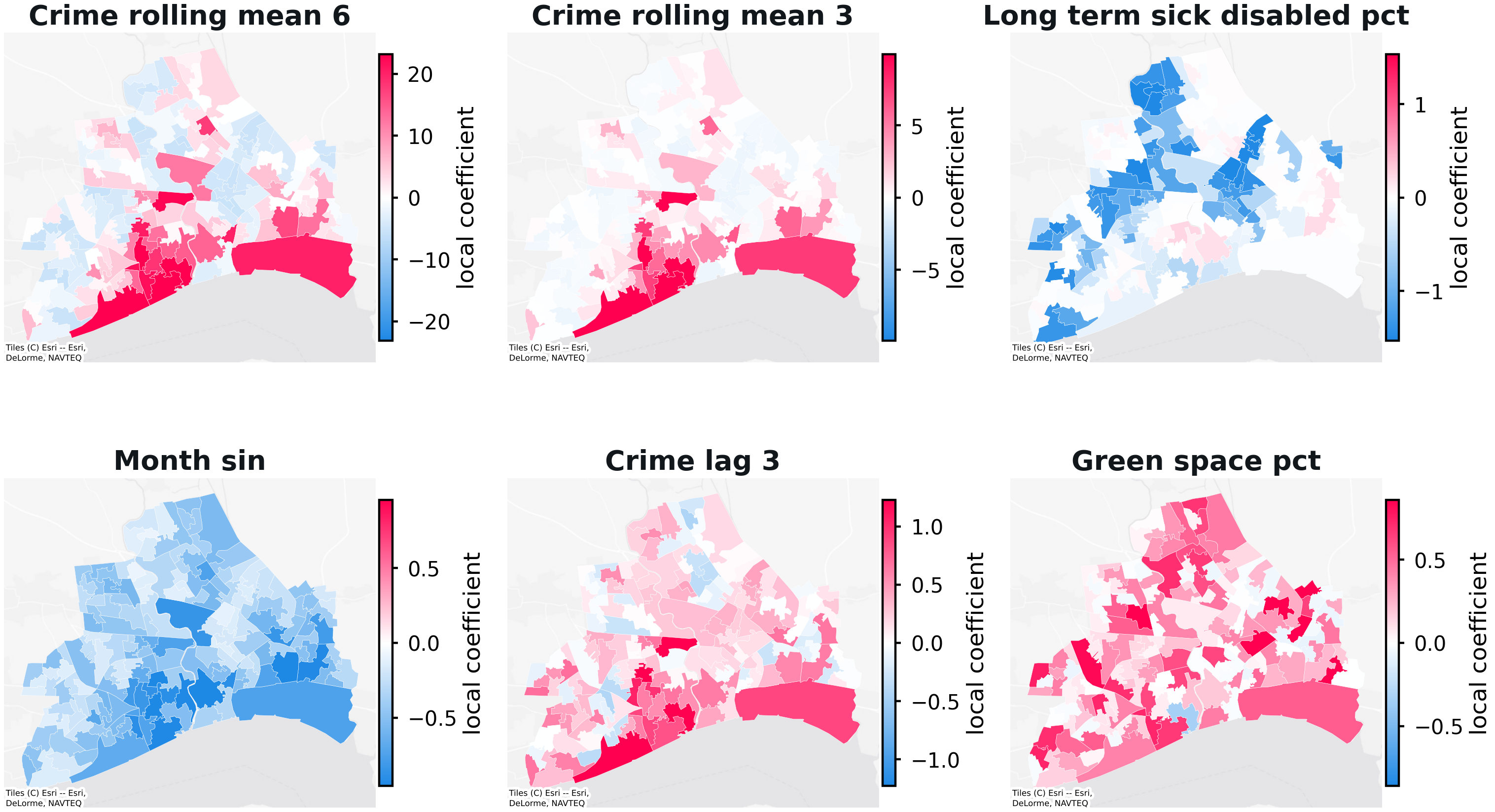}
\caption{Hull: signed local Feature coefficients, averaged over explained test months for each LSOA. Pink denotes a higher expected prediction when the predictor is retained relative to its median-masked reference; blue denotes a lower expected prediction. Per-panel colour scales differ.}\label{fig:hullfeatures}\end{figure}

Figure~\ref{fig:hullfeatures} shows the local Feature coefficients for Hull, including spatial variation in recent crime history and socioeconomic or built-environment predictors.

\begin{figure}[htbp]\centering
\includegraphics[width=.98\textwidth]{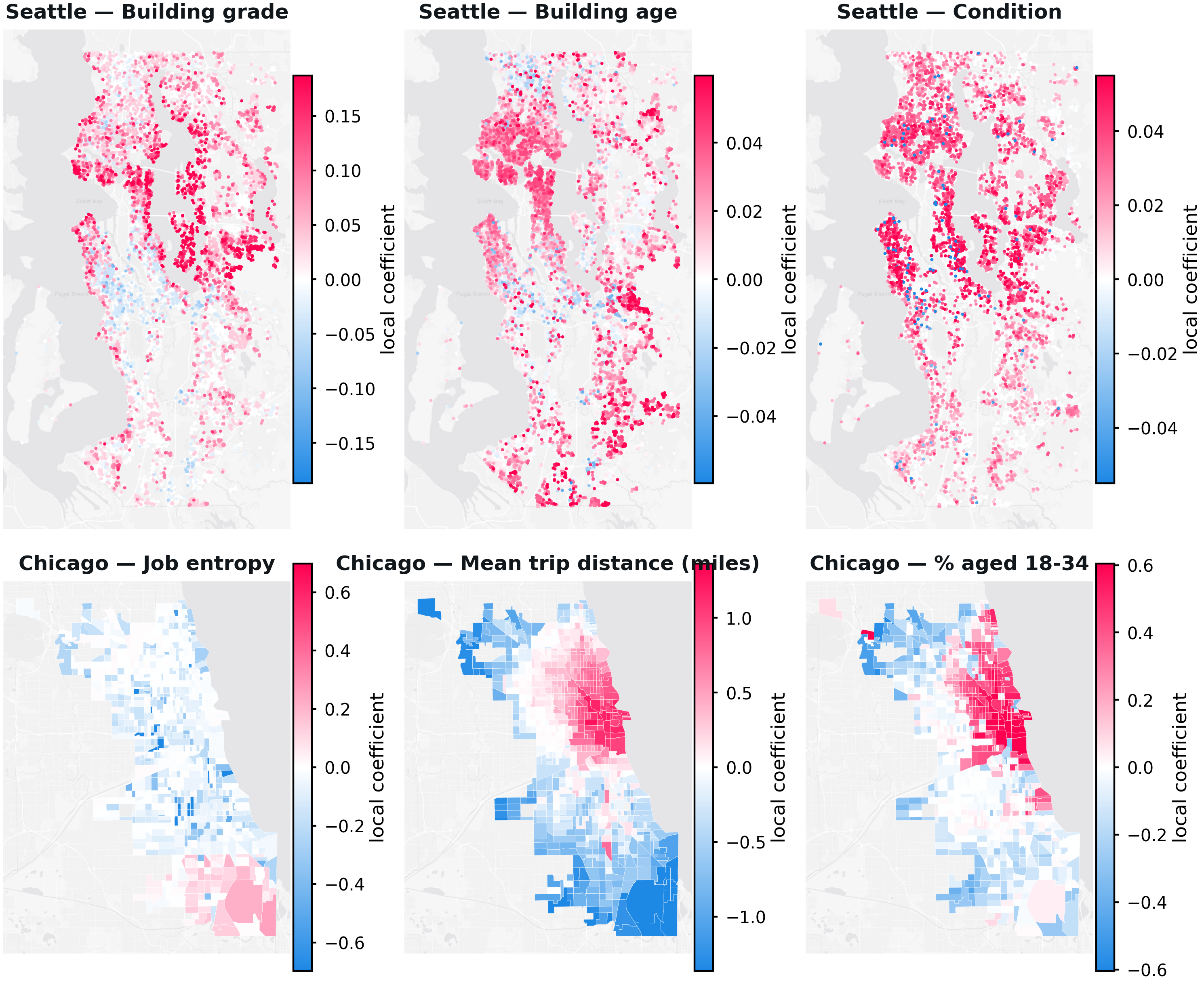}
\caption{Local Feature coefficients for Seattle (upper row) and Chicago (lower row). Signs indicate the expected prediction contrast associated with retaining the substantive predictor
}\label{fig:localmaps}\end{figure}

Figure~\ref{fig:localmaps} presents the Seattle and Chicago maps: construction grade, building age and condition vary across Seattle, while job entropy, mean trip distance and the share of young adults exhibit distinct patterns in Chicago. Together, these maps illustrate the spatial variation captured by the local Feature surrogates. 

\begin{figure}[htbp]\centering
\includegraphics[width=.98\textwidth, trim=0cm 0cm 0cm 0cm, clip]{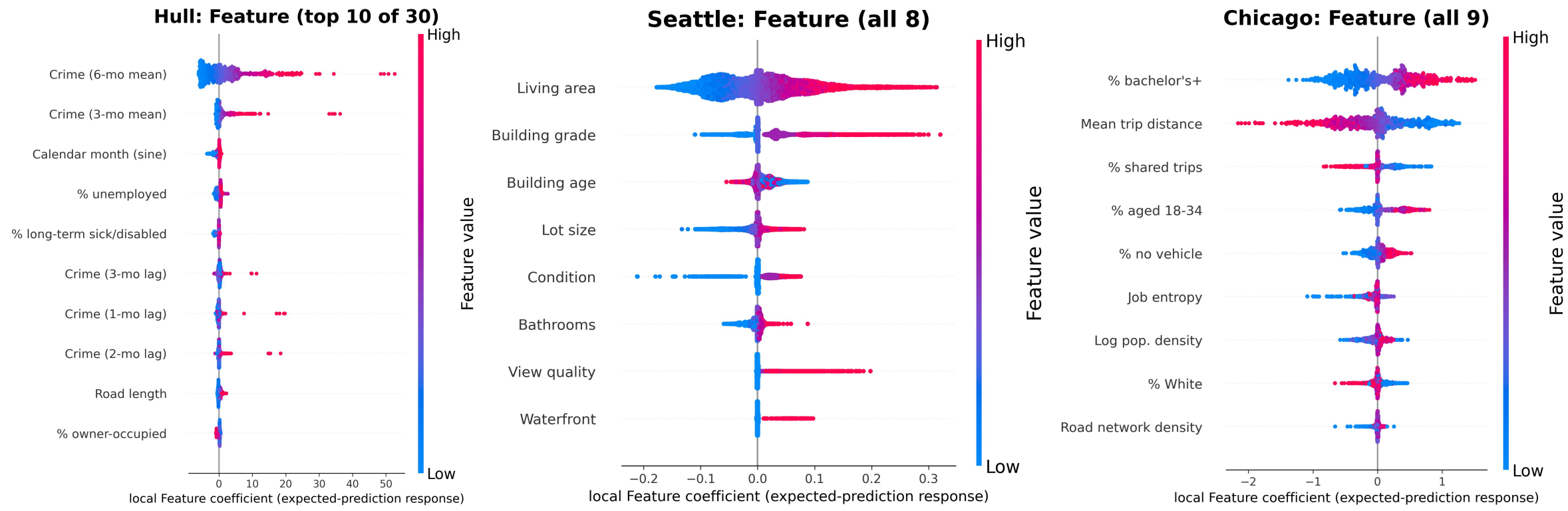}
\caption{Local Feature-coefficient distributions by predictor.
Each point is one explained observation; its colour is that predictor's own value (blue = low, pink = high)
}\label{fig:featurebeeswarm}\end{figure}

Figure~\ref{fig:featurebeeswarm} complements these maps with a non-spatial, per-predictor view: for each predictor, the distribution of local coefficients across all explained observations, coloured by the predictor's own value rather than by coefficient sign.

\FloatBarrier

\subsection{GeoSensitivity surfaces}
GeoSensitivity, defined in Equation~\eqref{eq:geosensitivity}, is signed in the keep orientation. The maps in this section display the magnitude of GeoSensitivity, $|\operatorname{GeoSensitivity}_g|$, whereas signed coefficients are retained for directional interpretation and full-vector consistency and stability comparisons; magnitude-based rankings are used for the top-ranked geographic-player analyses. Because the prediction target is averaged over a fixed observation support, unequal geographic-player group sizes can affect coefficient magnitudes. 

\begin{figure}[htbp]\centering
\includegraphics[width=.99\textwidth]{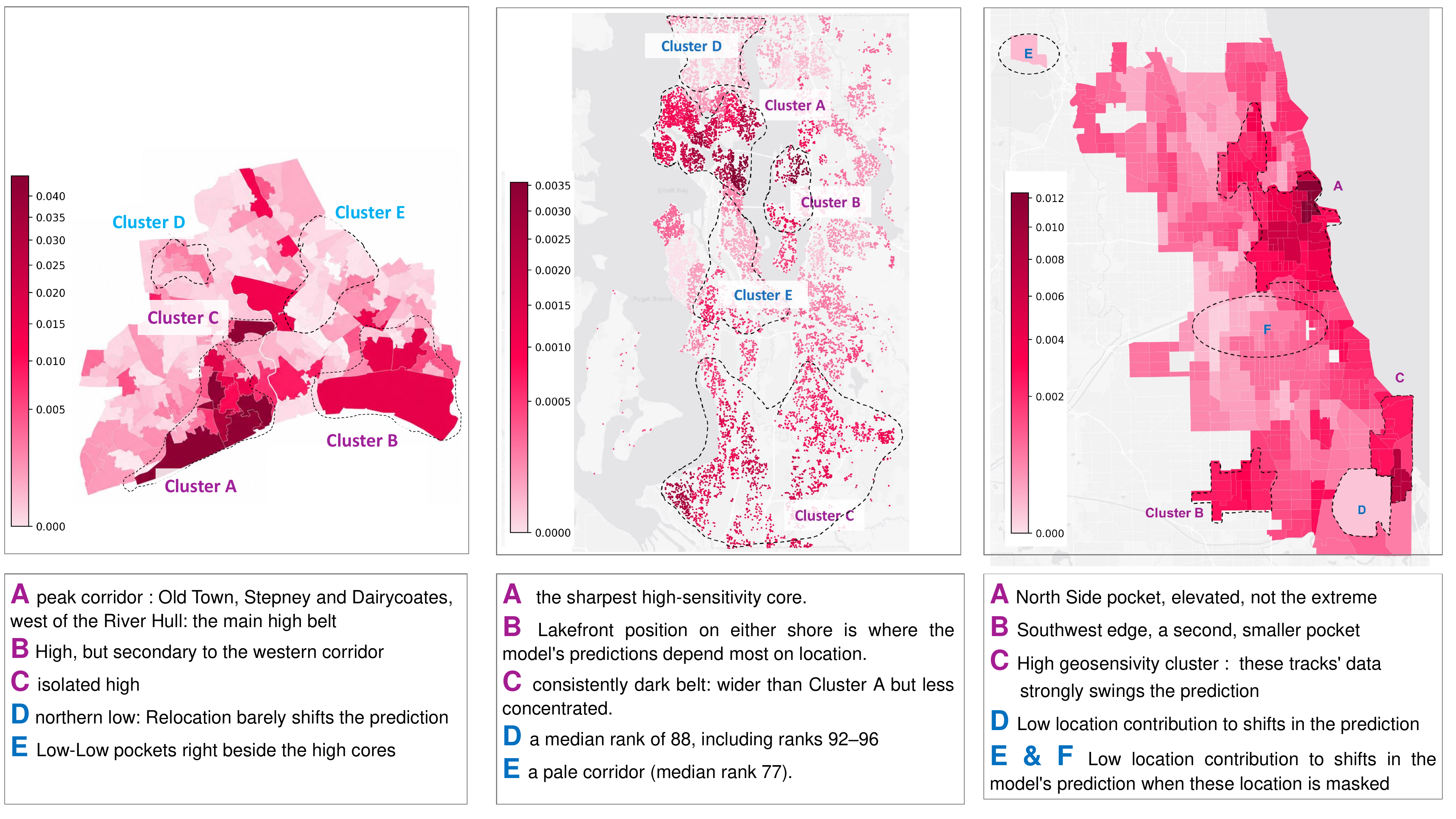}
\caption{Magnitude of GeoSensitivity, $|\widehat\beta_g^G|$, in Hull, Seattle and Chicago. Darker shading indicates larger geographic-player coefficient magnitudes under the specified relocation design.
Differences in response scale and grouping preclude direct cross-city magnitude comparisons.}\label{fig:geosurf}\end{figure}
\FloatBarrier

The resulting sensitivity estimates are defined for the fitted model, spatial grouping, donor operator and reference. The principal geographic supports are 168 LSOAs in Hull, 100 KMeans groups of house sales in Seattle and 160 KMeans groups of Chicago's 795 census tracts. Figure~\ref{fig:geosurf} maps this magnitude. For Seattle and Chicago, each observed location is coloured using the coefficient of its assigned geographic player; this does not imply an independently estimated coefficient at each sale or tract.

In Hull, larger magnitudes are concentrated in selected central and riverside LSOAs. Seattle shows concentrations among portions of the urban housing market, while Chicago exhibits contrasts across its grouped tracts. The maps depict the spatial distribution of coefficient magnitudes and are not formal tests of geographic clustering.

\subsection{ATT fidelity}
Table~\ref{tab:fidelity} reports test-set fidelity of the weighted Ridge surrogates fitted to expected black-box predictions under the Geo-SMILE perturbation design. The dependent variable is an expected black-box prediction, while locality weights are derived from summed marginal input-Wasserstein distances. The reported $R^2$ and weighted $R_w^2$ therefore measure approximation of this prediction target, rather than fidelity to a distribution-shift response. The test sets are held out from surrogate fitting; masks are separated across train, validation and test partitions according to the reported experimental protocol.

\begin{table}[htbp]\centering
\caption{Held-out surrogate fidelity for the Feature and Geo branches under the expected-black-box-prediction target.}\label{tab:fidelity}
\begin{tabular}{llrrrr}\toprule
Case & Branch & $R^2$ & $R_w^2$ & WMAE & WRMSE\\\midrule
Hull & Feature (30 players) & 0.906 & 0.910 & 0.3137 & 0.3702\\
Hull & Geo (168 players) & 0.903 & 0.906 & 0.0030 & 0.0040\\
Seattle & Feature (8 players) & 0.989 & 0.989 & 0.0011 & 0.0013\\
Seattle & Geo (100 players) & 0.886 & 0.842 & 0.0010 & 0.0012\\
Chicago & Feature (9 players) & 0.936 & 0.940 & 0.0135 & 0.0184\\
Chicago & Geo (160 players) & 0.850 & 0.844 & 0.0022 & 0.0028\\
\bottomrule
\end{tabular}\end{table}
\FloatBarrier

Across Hull, Seattle and Chicago, Feature-branch test weighted $R^2$ is 0.910, 0.989 and 0.940, respectively; the corresponding Geo-branch values are 0.906, 0.842 and 0.844. The Geo players comprise 168 Hull LSOAs, 100 Seattle point groups and 160 Chicago geographic groups in the evaluated configurations. These are branch-specific surrogates with different response scales and explanatory player sets; their coefficient magnitudes should not be compared directly across cases.

\subsection{ATT stability}
In Hull, stability was assessed by adding one real neighbouring LSOA in each cardinal direction (three LSOA-month rows), holding the fitted black-box model and reference fixed. Signed Geo Spearman ranges from $0.939$ to $0.944$, while $|\beta|$ Spearman ranges from $0.862$ to $0.871$. Magnitude-based top-20 Jaccard is $0.90$--$1.00$, retaining $95$--$100\%$ of the original top 20; none of the four added LSOAs enters the top 20. The Feature full-ranking Spearman is $0.974$ and the top-three predictors remain unchanged in every direction.

Chicago uses four quiet-anchor hold-outs, each withholding five tracts before retraining XGBoost and reintroducing that cluster as one new Geo player. Across directions, the signed Geo Spearman is $0.968$--$0.982$, magnitude-ranking Spearman is $0.906$--$0.944$, and top-20 magnitude retention is $85$--$100\%$. These tests reflect both support changes and black-box retraining, unlike the fixed-model Hull protocol.

\begin{figure}[H]
\centering
\includegraphics[width=\textwidth]{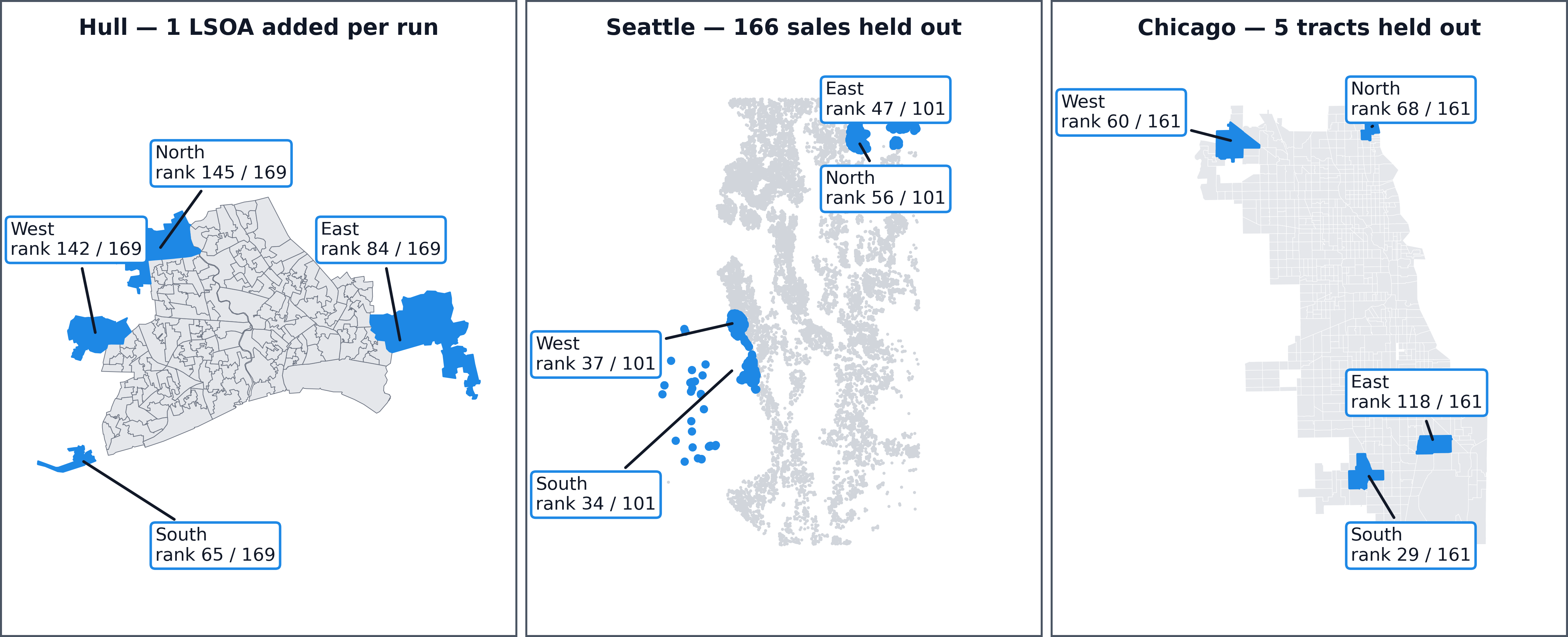}
\caption{Directional boundary-stability checks for the Feature and Geo branches in Hull, Seattle and Chicago.}
\label{fig:boundary-stability}
\end{figure}

Figure~\ref{fig:boundary-stability} illustrates the four directional boundary checks for both branches across the three cases.

Seattle Geo stability uses four quiet-anchor hold-outs of 166 sales each, with XGBoost retrained before each withheld cluster is added back as one new geographic group. Across north, south, east and west, the signed Geo Spearman ranges from $0.991$ to $0.994$ (mean $0.992$), and magnitude-ranking Spearman ranges from $0.973$ to $0.983$ (mean $0.979$). The magnitude-based top-20 Jaccard ranges from $0.74$ to $0.90$ (mean $0.840$), retaining $85$--$95\%$ of the original top 20 (mean $91.25\%$). The added groups rank $85$, $68$, $99$ and $92$ of $101$ (mean $86$); none enters the top 20. The Seattle Feature stability results show full-ranking Spearman and top-$k$ Jaccard both at $1.000$.

\begin{table}[H]\centering
\caption{Feature and Geo stability under directional boundary changes.}\label{tab:stability}
\begin{tabular}{llrrr}\toprule
Branch & Measure & Hull & Seattle & Chicago\\\midrule
Feature & Top-$k$ Jaccard & 1.000 & 1.000 & 1.000\\
Feature & Full-vector Spearman & 0.974 & 1.000 & 0.995\\
Geo & Top-20 Jaccard & 0.925 & 0.840 & 0.865\\
Geo & Full-vector Spearman & 0.942 & 0.992 & 0.978\\
Geo & Mean rank of added unit & 109 of 169 & 86 of 101 & 68.8 of 161\\\bottomrule
\end{tabular}\end{table}
Table~\ref{tab:stability} summarises the Hull, Seattle and Chicago tests. Each city contributes four directional tests. Geo top-20 Jaccard refers to magnitude-ranked coefficients, while Geo full-vector Spearman refers to the signed coefficients; the table reports four-direction means. Chicago Feature top-$k$ Jaccard concerns the top three features. Seattle and Chicago tests include model retraining, unlike Hull.
\FloatBarrier

\subsection{ATT consistency}
Consistency is summarised by the standard deviation of each fitted coefficient across five independently seeded repetitions, with the mean and maximum of these standard deviations reported over players. Hull and Chicago use seeds 42, 7, 123, 2024 and 31415; Seattle uses seeds 42, 52, 62, 72 and 82. Table~\ref{tab:consistency} reports the mean and maximum coefficient SD for each branch and case.

\begin{table}[htbp!]\centering
\caption{Five-seed coefficient consistency, summarised by mean and maximum coefficient standard deviation.}\label{tab:consistency}
\begin{tabular}{llrrr}\toprule
Branch & Statistic & Hull & Seattle & Chicago\\\midrule
Feature & Mean SD & 0.012272 & 0.00028 & 0.001565\\
Feature & Maximum SD & 0.023327 & 0.00040 & 0.002299\\
Geo & Mean SD & 0.000357 & 0.00349 & 0.000192\\
Geo & Maximum SD & 0.001368 & 0.01727 & 0.000471\\\bottomrule
\end{tabular}\end{table}
\FloatBarrier

\subsection{ATT accuracy against the Hull expert reference}
\begin{figure}[H]
\begin{minipage}[t]{.47\textwidth}
\vspace{0pt}\centering
\includegraphics[width=\textwidth]{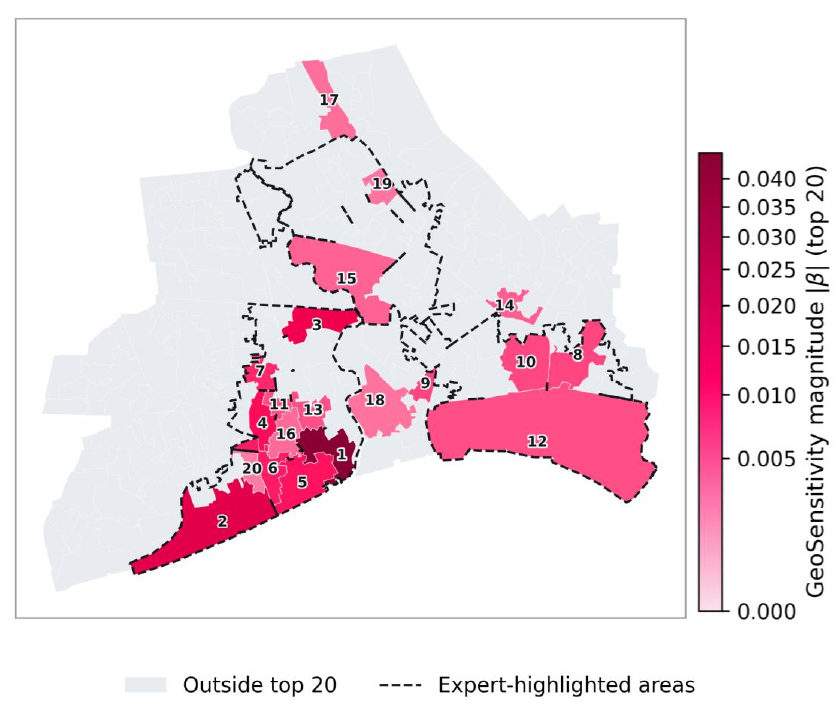}
\captionof{figure}{Hull: the 20 most sensitive LSOAs and the expert-highlighted areas.}\label{fig:expert}
\end{minipage}%
\hfill
\begin{minipage}[t]{.49\textwidth}
\vspace{0pt}\centering
\small
\begin{tabular}{lcc}
\toprule
Metric & Feature & Geo \\
\midrule
$n$             & 30     & 168 \\
Reference size  & 10     & 58 \\
$k$             & 10     & 20 \\
Intersection    & 7      & 16 \\
Top-$k$ accuracy & 0.7000 & 0.8000 \\
Reference prevalence & 0.3333 & 0.3452 \\
Lift            & 2.10   & 2.32 \\
Hypergeometric $p$ & 0.0048 & $1.25\times10^{-5}$ \\
ATT ACC         & 0.8000 & 0.7262 \\
ATT F1          & 0.7000 & 0.4103 \\
ATT AUROC       & 0.8450 & 0.6027 \\
Precision@$k$   & 0.7000 & 0.8000 \\
Recall@$k$      & 0.7000 & 0.2759 \\
Jaccard@$k$     & 0.5385 & 0.2581 \\
\bottomrule
\end{tabular}
\captionof{table}{Reference-based ATT accuracy in Hull at $N=4{,}000$: Feature top 10 versus 10 expert-named predictors, and Geo top 20 versus 58 expert-highlighted LSOAs.}
\label{tab:hull_att_accuracy_primary}
\end{minipage}
\end{figure}

Table~\ref{tab:hull_att_accuracy_primary} reports Hull ATT accuracy against the independently specified expert reference. Feature explanations recover 7 of the 10 reference predictors (precision@$10$ = 0.70; AUROC = 0.8450). Geo includes 16 expert-highlighted LSOAs among its 20 highest-ranked geographic players (precision@$20$ = 0.80; lift = 2.32; AUROC = 0.6027). Both overlaps are significant under the one-sided hypergeometric test of Equation~\eqref{eq:hypergeom} (Feature $p=0.0048$; Geo $p=1.25\times10^{-5}$). Figure~\ref{fig:expert} shows the geographic overlap. These measures answer different questions: AUROC summarises ranking separation across all 168 LSOAs under a reference that marks 58 of them (34.5\%, the table's reference prevalence) as expert-highlighted, whereas precision@$20$, lift and the reported overlap concern only the 20 highest-ranked units.
Because $k=20$ is smaller than the 58-unit reference, recall@$20$ is mechanically bounded above by $20/58\approx0.345$, so the observed $0.2759$ is close to that ceiling rather than indicating weak recovery.

\FloatBarrier

\subsection{Complementary diagnostics}
In what follows, we are testing sensitivity to the perturbation budget and model configuration to better understand the role of Perturbation in this study.

\subsubsection{Perturbation budget}\label{sec:budget}
The common coalition budget of $N=4{,}000$ was selected based on preliminary perturbation-budget experiments conducted during the underlying thesis research. Table~\ref{tab:budget} reports fidelity across seven coalition budgets, showing that $N=4{,}000$ provides the highest observed Geo fidelity in Hull and Chicago, while larger budgets do not consistently improve performance. The same budget is used across all three cases for consistency.
\begin{table}[htbp!]\centering
\caption{Held-out weighted surrogate fidelity ($R_w^2$) across perturbation budgets $N$. Seattle and Chicago Geo use 100 and 160 geographic groups, respectively.}\label{tab:budget}
\small
\setlength{\tabcolsep}{4pt}
\begin{tabular}{llrrrrrrr}\toprule
Case & Branch & 1,000 & 2,000 & 3,000 & 4,000 & 5,000 & 7,500 & 10,000\\\midrule
Hull & Feature & 0.911 & 0.905 & 0.909 & \textbf{0.910} & 0.909 & 0.907 & 0.903\\
Hull & Geo & 0.831 & 0.862 & 0.875 & \textbf{0.906} & 0.898 & 0.890 & 0.889\\
Seattle & Feature (8) & 0.989 & 0.988 & 0.988 & \textbf{0.989} & 0.987 & 0.988 & 0.988\\
Seattle & Geo (100, per-player) & 0.835 & 0.837 & 0.840 & \textbf{0.842} & 0.842 & 0.835 & 0.835\\
Chicago & Feature & 0.938 & 0.920 & 0.918 & \textbf{0.941} & 0.939 & 0.939 & 0.937\\
Chicago & Geo (160 groups) & 0.818 & 0.802 & 0.829 & \textbf{0.844} & 0.835 & 0.838 & 0.839\\\bottomrule
\end{tabular}\end{table}

\subsubsection{Design diagnostics}\label{sec:design}

Five interpretable linear estimators were compared to assess the sensitivity of surrogate fidelity to the regression method. Geographic grouping, donor selection and perturbation settings were held fixed across estimators.

Seattle's Feature-branch jitter treats its four binary or ordinal predictors (waterfront, view, grade, condition) the same as the continuous ones (Section~\ref{subsec:geosmile_formulation}). An additional experiment refits the same $N=4{,}000$ coalition masks with jitter disabled for these four predictors: held-out weighted $R^2$ is $0.982$, compared with $0.989$ under the primary configuration, and the eight fitted coefficients agree closely between configurations (Pearson $r=0.98$ for signed coefficients; Spearman $\rho=0.93$ for absolute coefficient rankings). The four highest-ranked predictors --- grade, age, condition and lot size --- retain their signs and ranks, indicating that the principal Seattle Feature attributions show limited sensitivity to the removal of Gaussian jitter from these predictors.

Surrogate family: Table~\ref{tab:surrogates} reports the held-out weighted $R_w^2$ results for five linear surrogates under the specified Geo-SMILE configurations. Across the five estimators, the Feature-branch scores range from 0.9096 to 0.9098 in Hull, from 0.9881 to 0.9885 in Seattle, and from 0.9356 to 0.9406 in Chicago. Geo-branch scores range from 0.9055 to 0.9076 in Hull, from 0.8344 to 0.8423 in Seattle, and from 0.8406 to 0.8444 in Chicago. The five linear estimators have similar fidelity, although Weighted Ridge is not the highest-scoring option in every case. It is retained for coefficient regularisation and a consistent estimation protocol. Chicago uses 160 geographic players.

\begin{table}[htbp!]\centering
\caption{Held-out weighted $R_w^2$ by linear surrogate model and case.}\label{tab:surrogates}
\begin{tabular}{lrrrrrr}\toprule
& \multicolumn{2}{c}{Hull} & \multicolumn{2}{c}{Seattle} & \multicolumn{2}{c}{Chicago}\\
Surrogate & Feature & Geo & Feature & Geo & Feature & Geo\\\midrule
\textbf{Weighted Ridge (proposed)} & \textbf{0.9098} & \textbf{0.9055} & \textbf{0.9885} & \textbf{0.8423} & \textbf{0.9405} & \textbf{0.8437}\\
Weighted OLS & 0.9097 & 0.9064 & 0.9885 & 0.8414 & 0.9406 & 0.8444\\
Bayesian ridge & 0.9097 & 0.9064 & 0.9885 & 0.8417 & 0.9406 & 0.8441\\
Linear regression (OLS) & 0.9096 & 0.9076 & 0.9881 & 0.8344 & 0.9356 & 0.8406\\
Ridge (unweighted) & 0.9096 & 0.9074 & 0.9881 & 0.8353 & 0.9356 & 0.8406\\\bottomrule
\end{tabular}\end{table}
\FloatBarrier
\section{Discussion}\label{sec:discussion}
\subsection{Geographic position as an explanatory target}
Geo-SMILE makes geographic position an explicit target of post-hoc explanation alongside the substantive predictors used by a fitted model. The two branches address distinct questions about the same fitted model: Feature explanations describe responses to substantive predictor masking, whereas GeoSensitivity describes responses to geographic relocation. Their separation prevents positional sensitivity from being conflated with ordinary coordinate-feature importance.

Because a relocated observation may combine coordinates and substantive attributes rarely observed in the original data, the fitted coefficients must be interpreted within the stated experimental design rather than as unconditional effects.

\subsection{What GeoSensitivity reveals across spatial supports}\label{sec:geosensitivity}
The magnitude patterns mapped in Figure~\ref{fig:geosurf} are shaped jointly by geographic support, donor selection and covariate configuration, not by relocation alone.

The three applications demonstrate how the same explanatory concept can be represented at different spatial supports. Hull uses 168 LSOAs as geographic players, so the mapped coefficients correspond directly to those areal units. Seattle uses 100 groups of housing transactions, while Chicago uses 160 groups formed from 795 tracts. In the latter two cases the map assigns each observation its group's coefficient; visually detailed maps therefore do not imply that a separate GeoSensitivity coefficient was estimated for every sale or tract. The patterns can be interpreted within each case, but raw coefficient magnitudes cannot establish that one city's predictive model is more geographically sensitive than another's, given differences in outcome scales, spatial support and perturbation context.

Local Feature maps add another layer of interpretation. Recent crime history and selected contextual predictors vary across Hull LSOAs; construction grade, age and condition display different local associations in Seattle; and job entropy, mean trip distance and the young-adult share vary across Chicago's tracts. These patterns answer a different question from GeoSensitivity: they concern the model's response to substantive attributes rather than its response to position. Reading the two outputs together can distinguish areas with strong location-conditioned behaviour from areas where particular predictors have locally varying fitted associations.

\subsection{Evidence for explanation reliability}
Geo fidelity is lower than Feature fidelity in Seattle and Chicago (Table~\ref{tab:fidelity}), indicating that geographic relocation responses are less closely approximated by the linear surrogate. This does not necessarily imply that the corresponding geographic coefficients are incorrect. The closely spaced scores across the tested linear estimators (Table~\ref{tab:surrogates}) suggest that this is not an artefact of the particular choice of weighted Ridge.

Stability and consistency examine different sources of variation: the five-seed coefficient standard deviations (Table~\ref{tab:consistency}) indicate reproducibility under repeated sampling, whereas the directional boundary tests (Table~\ref{tab:stability}) probe changes to the geographic player support. Because Hull's protocol retains the fitted black box while Seattle's and Chicago's retrain it, these results support within-case robustness assessments rather than a controlled ranking of the cities by stability. The Hull reference-set overlap provides complementary evidence for reference-based attribution accuracy.

The perturbation-budget results (Table~\ref{tab:budget}) show that increasing the coalition budget beyond $N=4{,}000$ did not consistently improve held-out Geo fidelity in the evaluated configurations: Hull and Chicago Geo fidelity peaks at $N=4{,}000$, with no consistent gain beyond it. Together, these diagnostics distinguish numerical agreement with the surrogate target from sensitivity to sampling choices and robustness to changes in spatial support.

\section{Limitations}\label{sec:limitations}
GeoSensitivity is conditional on geographic grouping, the number of observations represented by each player and the donor distribution. The coefficients therefore describe the fitted model under the stated
intervention, not causal location effects or a correction for spatial autocorrelation.  

Ground-truth geographic attributions are generally unavailable in real-world datasets. Accordingly,
Geo-SMILE is evaluated through complementary measures of fidelity, stability and consistency, alongside
an independent expert reference in Hull. These assessments provide evidence of explanation quality and
reproducibility across the three applications.

Although model-agnostic in design, Geo-SMILE was evaluated using XGBoost across three geospatial
applications. Further evaluation with other model families would help establish its broader generalisability.

\section{Future research}\label{sec:future}
Future research should assess sensitivity to player grouping, group size, spatial resolution and donor selection, including relocation constraints that reduce unsupported attribute--coordinate combinations. Spatially informed multivariate distances could then be evaluated against the marginal Wasserstein kernel used in this study while retaining the same explanation target.

Controlled spatial benchmarks with known model-response functions would help test recovery of intervention-conditioned sensitivity. Additional work should use comparable explanatory targets when benchmarking against spatial XAI methods, and repeat the existing stability protocols under matched conditions. Evaluation across other predictive model families and datasets would clarify where a simple linear surrogate adequately represents complex geographic responses.

\section{Conclusion}\label{sec:conclusion}
This study introduced Geo-SMILE to explain geographic-position sensitivity alongside substantive-feature associations in geospatial prediction models. Its main contribution is GeoSensitivity, a geographic-player surrogate coefficient, signed in the keep orientation, under a specified location-relocation design. By separating the Feature and Geo branches, the framework gives geographic position an explicit explanatory role alongside substantive predictors, with each branch addressing a distinct model-response question.

The formulation adapts SMILE's statistical-distance-based surrogate weighting to geographic data. Feature masking retains the observed coordinates, whereas geographic-player relocation retains substantive attributes; both branches fit weighted linear surrogates to expected black-box predictions. The same formulation was applied to Hull crime, Seattle house prices and Chicago ride-hailing demand using spatial units appropriate to each dataset. Held-out weighted fidelity ranged from $0.910$ to $0.989$ for Feature and from $0.842$ to $0.906$ for Geo. The mapped coefficients show variation in the model's conditional responses, while the evaluation distinguishes surrogate approximation, reproducibility and expert-reference agreement.

The results establish the feasibility of interpreting substantive and geographic dimensions through a common perturbation-surrogate framework, while also identifying boundaries on what such explanations mean. GeoSensitivity depends on the spatial support, donor operator and fitted predictor. The accompanying stakeholder dashboard illustrates how the resulting global and local explanations can be communicated to audiences with different analytical needs. Further tests across spatial supports, relocation schemes and model families will be important for establishing the wider applicability of the approach.

\section*{Data and code availability}
The Seattle and Chicago datasets follow the sources used in the corresponding spatial-XAI studies \parencite{li2024,li2022}.

Hull crime data come from data.police.uk \parencite{homeoffice2026}; census and OpenStreetMap data sources are documented in the paper and the supporting materials. 

The implementation, the result files behind the reported tables and figures, setup instructions and data-source documentation will be made available at \url{https://github.com/shadiehh/Geo-SMILE} upon publication. Access to particular raw datasets remains subject to their source licences and terms.

\section*{List of Abbreviations}
\begin{longtable}{l L{\dimexpr\textwidth-3cm\relax}}
\toprule
Abbreviation & Definition\\\midrule
\endhead
\bottomrule
\endlastfoot
ACC & Accuracy\\
ANN & Artificial Neural Network\\
ATT & Attribution\\
AUROC & Area Under the Receiver Operating Characteristic Curve\\
BNG & British National Grid\\
CNN & Convolutional Neural Network\\
ECDF & Empirical Cumulative Distribution Function\\
EPA & U.S. Environmental Protection Agency\\
EPSG & European Petroleum Survey Group (coordinate reference system code)\\
GeoAI & Geospatial Artificial Intelligence\\
Geo-SMILE & Geospatial extension of SMILE proposed in this paper\\
GeoXAI & Geospatial Explainable Artificial Intelligence\\
GeoXCP & Geospatial Explanation with Conformal Prediction\\
GLIME & Geographically Localised Interpretable Model-Agnostic Explanations\\
GPI & Geographical Pattern Interaction model\\
GWR & Geographically Weighted Regression\\
LIME & Local Interpretable Model-Agnostic Explanations\\
LSOA & Lower Layer Super Output Area\\
MAE & Mean Absolute Error\\
MLP & Multilayer Perceptron\\
OLS & Ordinary Least Squares\\
RBF & Radial Basis Function\\
RMSE & Root Mean Squared Error\\
SA-SHAP & Spatially Adjusted SHAP\\
SD & Standard Deviation\\
SHAP & SHapley Additive exPlanations\\
SMILE & Statistical Model-Agnostic Interpretability with Local Explanations\\
SVR & Support Vector Regression\\
UTM & Universal Transverse Mercator\\
WMAE & Weighted Mean Absolute Error\\
WRMSE & Weighted Root Mean Squared Error\\
XAI & Explainable Artificial Intelligence\\
\end{longtable}
\FloatBarrier
\printbibliography
\end{document}